\documentclass[sigconf,authorversion,nonacm]{acmart}

\AtBeginDocument{%
  }

\setcopyright{none}
\begin{document}
\pagestyle{plain}

\title[Emotion Recognition and Forecasting in Conversation (ERFC)]{ERFC: Happy Customers with Emotion Recognition and Forecasting in Conversation in Call Centers}

\author[Aditi Debsharma et al.]{Aditi Debsharma,
        Bhushan Jagyasi, 
        Surajit Sen,
        Priyanka Pandey,\\
        Devicharith Dovari, 
        Yuvaraj V.C, 
        Rosalin Parida, 
        Gopali Contractor}
        
\affiliation{
{
\institution{Center for Advanced AI, Accenture} 
\country{}}
\{{aditi.debsharma}, {bhushan.jagyasi}, {surajit.sen}, {p.l.pandey}, {devicharith.dovari}, {yuvaraj.v.c}, {rosalin.parida}, {gopali.contractor}\}@accenture.com}

\begin{abstract}
  Emotion Recognition in Conversation (ERC) has been seen to be widely applicable in call-center analytics, opinion mining, finance, retail, healthcare, and other industries. In a call-center scenario, the role of the call center agent is not just confined to receiving calls but to also provide good customer experience by pacifying the frustration or anger of the customers. This can be achieved by maintaining neutral and positive emotion from the agent’s end. As in any conversation, the emotion of one speaker is usually dependent on the other speaker’s emotion hence the agent’s positive emotion accompanied with the right resolution will help in enhancing customer experience. This can change an unhappy customer to a happy one. Imparting the right resolution at right time becomes easier if the agent has the insight of the emotion of future utterances. To predict the emotions of the future utterances we propose a novel architecture \emph{Emotion Recognition and Forecasting in Conversation (ERFC)}. Our proposed ERFC architecture considers multi-modalities, different attributes of emotion, context and the inter-dependencies of the utterances of the speakers in the conversation. Our intensive experiments on the IEMOCAP dataset have shown the feasibility of the proposed ERFC. This approach can provide a tremendous business value for the applications like call center, where the happiness of customer is utmost important.
\end{abstract}


\begin{teaserfigure}
\centering
  \includegraphics[width=0.65\textwidth]{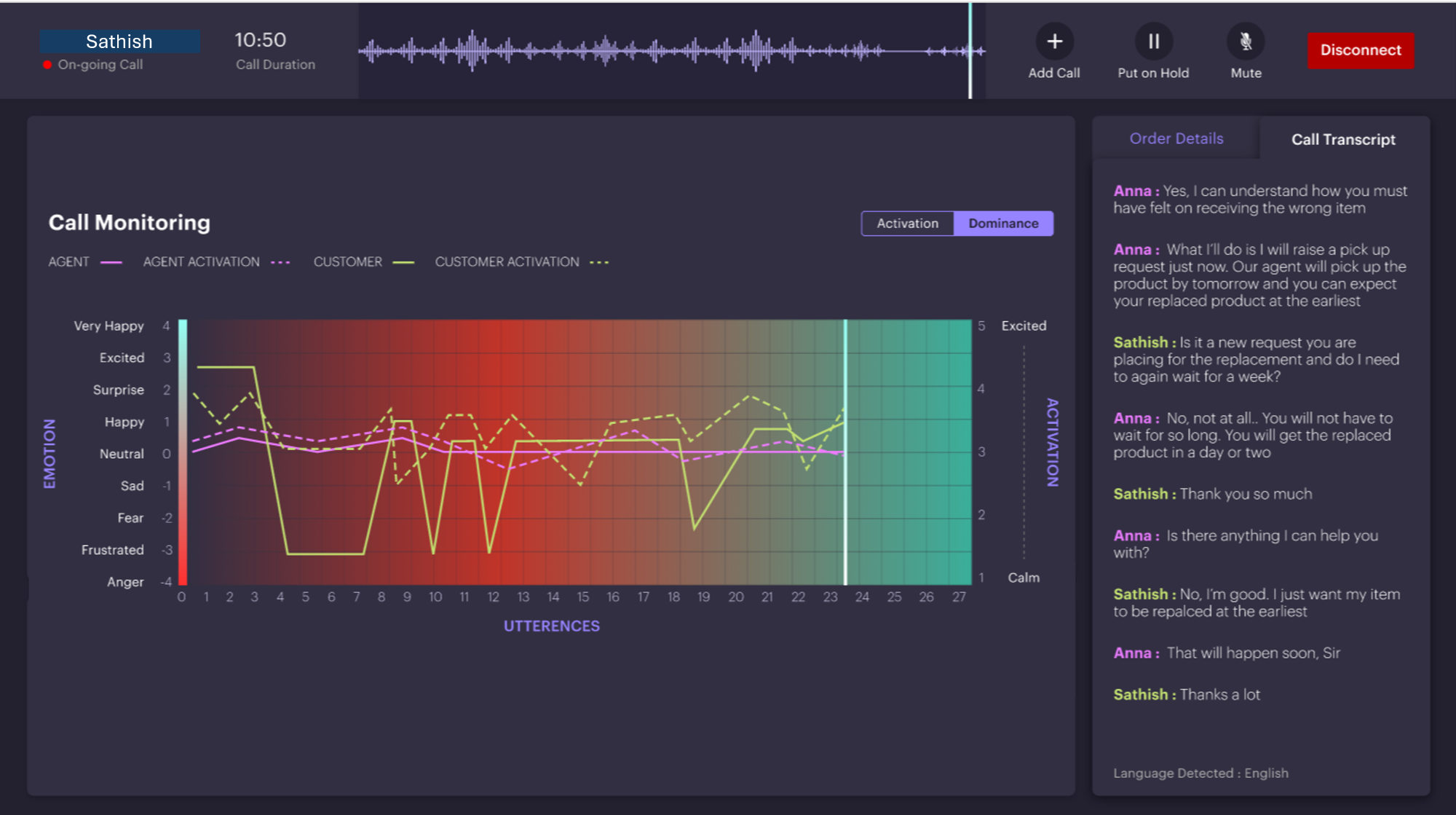}
  \caption{ERFC in Agent's console in Call-center}
\end{teaserfigure}

\maketitle

\section{Introduction}
In any conversation, emotions expressed are highly inter-dependent on the emotions of the other speakers involved. 
A change in the emotion of one speaker might lead to the change in the emotion of the other speaker over the subsequent utterances. 
For example, mostly, a consistent positive emotion of one speaker can bring down the intensity of the negative emotion of the other speaker in a conversation and vice versa. 
Similarly, in any call center there are different emotions observed during conversation with the customer. 
However, the agent have to maintain a positive emotion irrespective of the emotion of the customer to enhance the customer experience. 
Typically, an agent should aim to convert an \emph{unhappy customer} towards a \emph{happy customer} with the right resolutions along with exhibiting the positive emotions. 
So, if an agent has an information about the future emotions of the customer, which is going towards the negative direction, it becomes easier for the agent to take corrective actions. 
In such a scenario, the agent can amplify their positive emotions to empathize towards an unhappy customer, in order to resolve the concerns in a right way. 

With the increasing demand of the application of emotion recognition, there have been significant developments in this area from emotion recognition for an utterance (without context) to Emotion Recognition in Conversation (ERC) where context from previous utterances are also taken into account.

In literature, for Emotion Recognition in Conversation (ERC), there have been multiple studies to enhance the state-of-the-art accuracy. 
Most of the studies in ERC have been accomplished by considering context and multimodal features \cite{DialogueTRM, DialogueGCN, DialogueRNN, lietal2020hitrans, poria2019review, tripathi2018multi} which is the main difference between ERC and Emotion recognition in utterance. These studies mainly focus on designing the right architecture to achieve higher accuracy. In our paper \emph{Varta Rasa} 
\cite{VartaRasa2023}, we have considered context to predict the emotion of the present utterance, however, with a light weight architecture to achieve a comparable accuracy.

Self dependencies of speaker for a context has been explored in \cite{bcLSTM, ICON, hazarika2018b}. 
In \cite{mutualinfluence}, the interdependencies between two consecutive \emph{turns} of utterance between the two speakers have been modeled as Dynamic Bayesian Network Structure for emotion Recognition. 
When two speakers in a conversation complete their utterance, that defines as one turn.

Lately, there have been some work which highlights the importance of emotion forecasting. In \cite{Audio-Visual-Emotion-Forecasting}, audio-visual cues are used to forecast the emotion of the next two turns for a single speaker. In the similar lines, there have been some other works which identified the importance of emotion forecasting in different use cases. The authors of \cite{EmotionPre-Recognition, predictingevokedemotionsconversations} highlights on the importance of emotion forecasting in human-machine interactions where pre-recognition of user's emotion can result in quality and successful conversation.

In this paper, we identify a gap in solutions available in literature for call center conversation application where agents do not get enough insights from the emotions of the current utterance. 
We hence see this scenario in a new light, by focusing on the \emph{emotion recognition} for the current utterance along with \emph{emotion forecasting} for the future utterances of the next few turns of both the speakers involved in the conversation. 
We propose Emotion Recognition and Forecasting in Conversation (ERFC) which fills the gap in the literature and emphasises the utility of this solution for business problems like customer retention in a call center setup. 
On the proposed ERFC, from a given conversation, apart from emotion recognition for the current utterance, the emotions for the future utterances over a few turns of conversation are forecasted. 

In ERFC, we have considered context, inter-dependency of emotion between speakers in conversation and turns. 
In addition to this we have also considered the attributal features of emotion which are - Activation, Valence, and Dominance (AVD) \cite{Busso2008}.  
Activation indicates speaker's excitement level which can vary between excited to calm, Valence is the tone of the speech and it can range between positive and negative, and Dominance is the strength of the voice which varies from strong to weak. 
The primary reason for considering attributal features along with emotions is that the latter cannot provide information on the intensity of the emotion. 

For the call center application, we aimed to forecast the next few turns of both the speakers. This helps the call center agent to get the insights of the future trend of the conversation.

Overall, the key contributions in this paper are:
\begin{itemize}
	\item We propose the Emotion Recognition and Forecasting in Conversation (ERFC) which is designed to be used for the applications such as call-center conversations for current utterance emotion prediction along with emotion forecasts of future utterances, for both the speakers.
	\item In the proposed ERFC, we utilize (a) multi-modal features (audio and text) (b) all three dimensional (attribute) features-valance, activation and dominance (AVD), of emotions to complement emotion and (c) mutual influence of the speakers in conversation by including speaker turns along with speaker information. 
	\item We propose a light weight multi-level stacking architecture which is based on our earlier work \emph{Varta Rasa} 
  \cite{VartaRasa2023} which was also a light weight architecture, albeit for ERC .	
	\item We have also done an exhaustive experimental analysis using Interactive Emotional Dyadic Motion Capture (IEMOCAP) corpus \cite{Busso2008} to present the results of our proposed ERFC solution. 	 
\end{itemize}

The remaining part of the paper is structured as follows: 
In the next section, the details of the proposed ERFC architecture has been presented; 
in section \emph{Experiments and Results}, the Experimental setting and Results of the implementation of ERFC using the IEMOCAP dataset have been presented; 
in section \emph{Business Impact}, the value realization of the proposed ERFC for the customer care support application and the other potential applications have been discussed; 
we finally conclude our work in the section \emph{Conclusion}.  

\section{The Proposed Emotion Recognition and Forecasting in Conversation (ERFC) Architecture}
\label{sec:Architecture}
In a conversation of two speakers, the emotion of one speaker plays an important role in influencing the emotion of the other speaker. 
Let's understand this with an example from the Interactive Emotional Dyadic Motion Capture (IEMOCAP) corpus \cite{Busso2008}. 
\begin{figure*}[!h]
	\centering
	\includegraphics[width=0.75\linewidth]{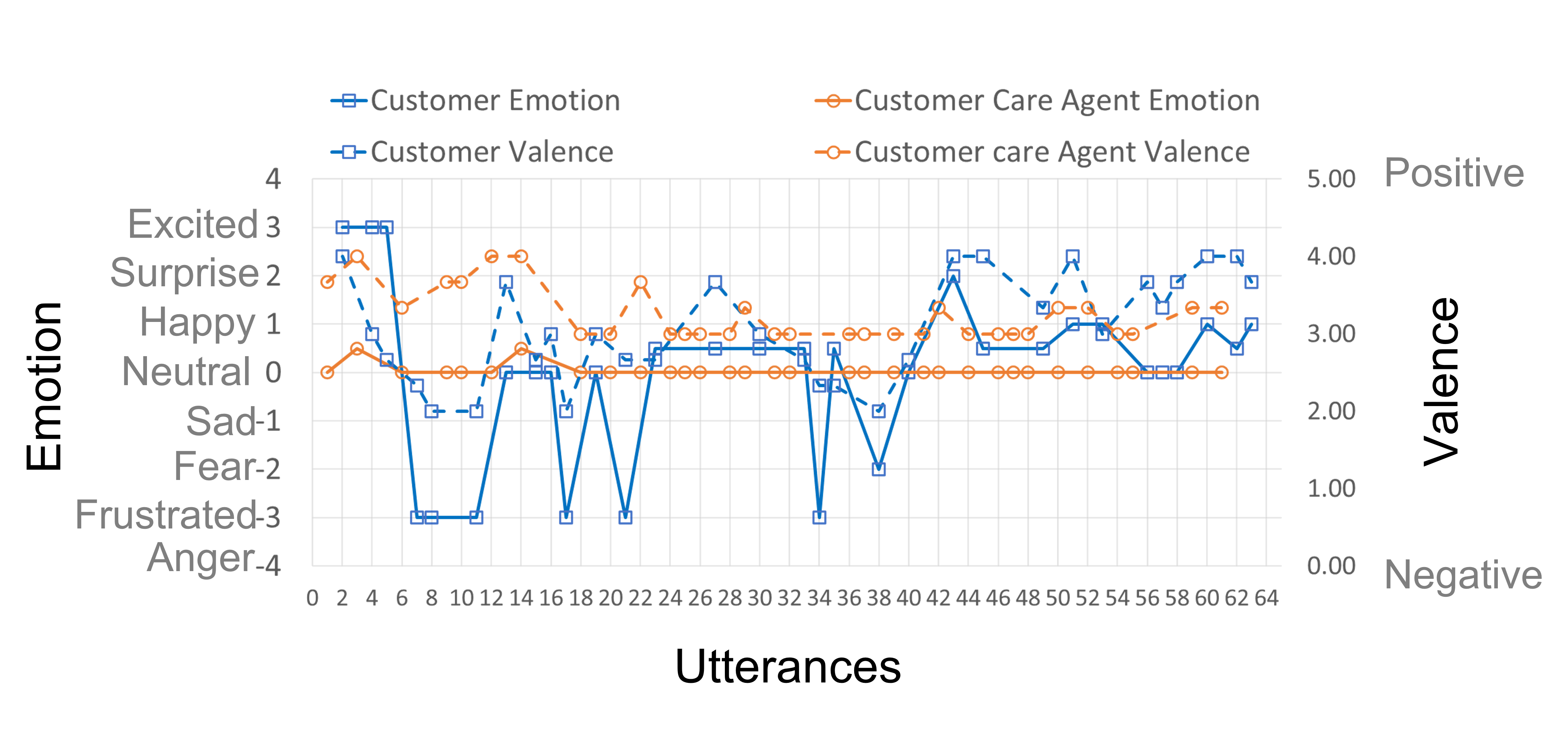}
	\caption{Mutual interdependence between two speakers from a conversation from IEMOCAP dataset}
	\label{fig:mutual_dependency}
\end{figure*}
Fig. \ref{fig:mutual_dependency} depicts emotion and valence expressed in a conversation between a customer and a customer care agent where the customer is seen to be excited in her initial utterances due to reaching out to a human agent which she has been trying for long on the IVR (Interactive Voice Response) call. 
Immediately after that, during utterances 7-11, the customer is seen to be frustrated due to the fact that she has been constantly re-directed to IVR which could not resolve her issue. 
The customer care agent understands the customer’s frustration and tries to give her the solution to avoid IVR in future with his consistent neutral emotion and positivity in his tone indicated by the valance. 
This provided the customer temporary satisfaction indicated by neutral emotion during the utterance 13-16. 
However, the customer’s primary concern and purpose of contacting the customer care agent is for some incorrect transaction in her card. 
While expressing and registering this complaint the customer is very anxious and her emotion again goes down towards the negative direction along with her valence during 17-26 utterance with few fluctuations. The agent understands the customer’s concern and endeavours to provide the best experience by giving proper solution with his consistent neutral emotion and positivity in the tone. 
The customer was bit skeptical with the initial solution, however the agent’s positive approach and tone reflected through his emotion and valence helped the customer gain some confidence and finally the conversation ends with a happy note reflected through emotion and valence in utterance between 51 to 63. 
With this example we observe that the consistent positive emotion of one speaker can influence the other speaker in the conversation positively. 
Similarly we have also observed in the other examples where a negative tone of agent has impacted the emotion negatively leading to an un-happy customer. 

\begin{figure*}[!h]
	\centering
	\includegraphics[scale=.65]{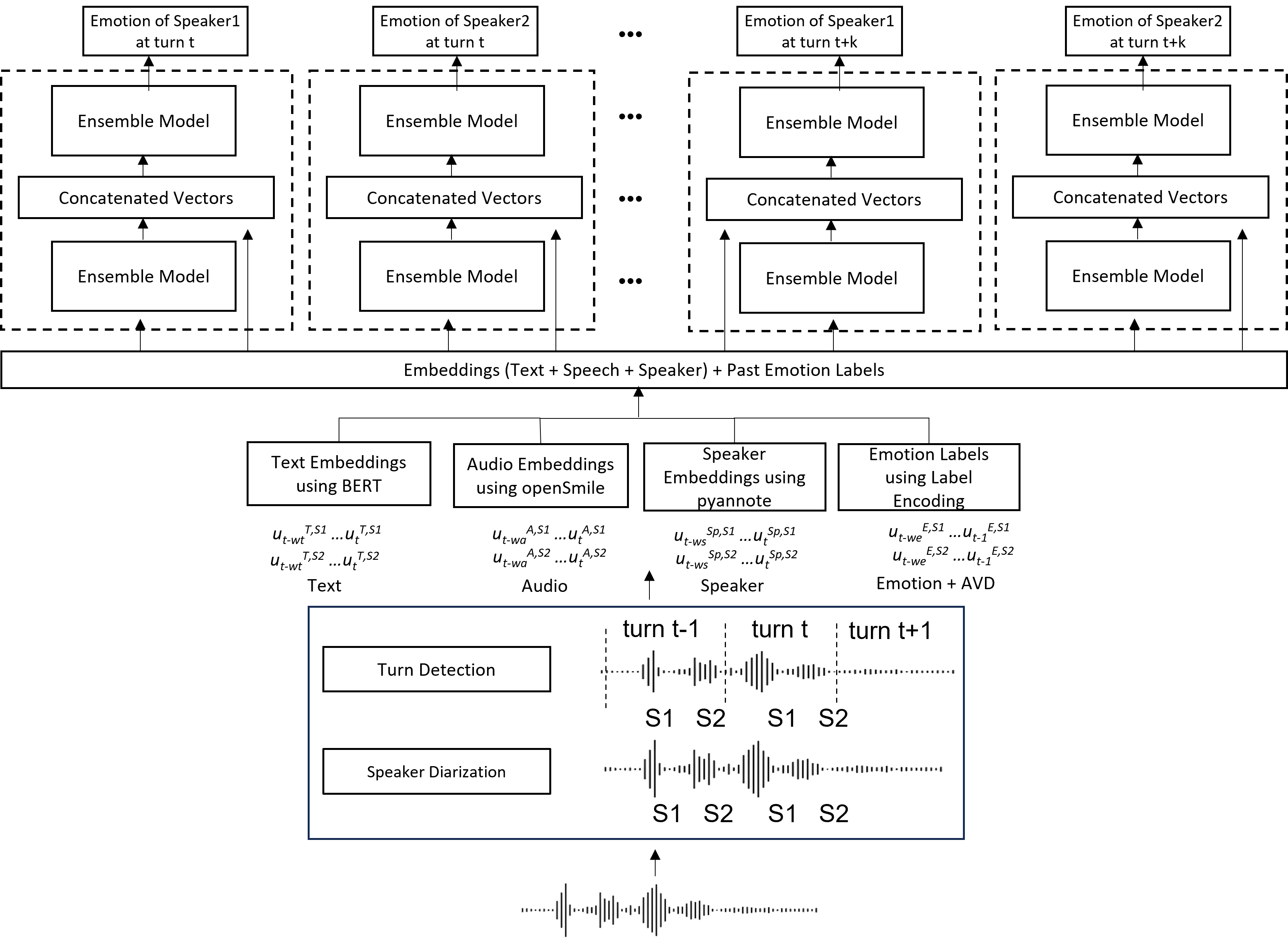}
	\caption{Proposed Emotion Recognition and Forecasting in Conversation (ERFC) architecture}
	\label{fig:arch}
\end{figure*}

In this paper we propose an Emotion Recognition and Forecasting in Conversation (ERFC) solution which leverages the inter-dependencies between the speakers to not only predict the emotion for the current utterance but also forecast the emotion for the future utterances. Fig. \ref{fig:arch} presents the architecture diagram of the ERFC solution which is based on inter-dependencies between the speakers, context window of past utterances and multi-Model features. The significance of these are explained as below: 
\begin{itemize}
\item To capture the mutual influence or inter-dependencies of emotion of one speaker over the emotion of the other, we consider the detection of turn (as in \cite{mutualinfluence}) by utilizing speaker identification through diarization in the conversation.
\item To predict the current emotion and forecast the future emotions more accurately, it is important to understand the trend of the emotions from the past utterances; we hence consider the context window as well.  
\item Multi model features relevant in a call center scenario are considered like audio, text, speaker information and emotion along with their attributal features of the past utterances. We have not considered video as this is not relevant in a call center scenario. 
\end{itemize}

Fig. \ref{fig:arch} shows that audio goes as input which then is passed through speaker diarization to extract the speaker information in terms of number of speakers involved in the conversation and their corresponding time stamps with respect to each utterance of the speaker. 
For example, in a given conversation where [$u_1$, $u_2$, $u_3$, ...., $u_N$] are the utterances, speaker diarization identifies the speaker with their corresponding utterance along with their timestamp. 
The speaker information is further leveraged to label \emph{turns} in the conversation. A single \emph{turns} is when both the speakers in a two speaker conversation complete their utterances. 

In our proposed ERFC, we have used multiple feature modalities and computed their respective embeddings to be used as input to the model. 
For text embeddings, we use pre-trained BERT language model from hugging face transformer library \cite{wolf2019huggingface, VartaRasa2023}; for audio embeddings, we use openSMILE \cite{Opensmile, VartaRasa2023}; for speaker embeddings, we use Pyannote \cite{Pyannote} and for emotions labeling, we use label encoding. Here emotion represent emotion and emotion attributes - activation, valence, and dominance.  
Embedding in all the modalities are created separately for both the speakers ($S_i$ where $i\in\{1, 2\}$). 
Here $u_t^{T, S_i}$, $u_t^{A, S_i}$, $u_t^{S_p, S_i}$ and $u_t^{E, S_i}$ represent text embeddings, audio embeddings, speaker embeddings, and emotion label for speaker $S_i$'s utterance in  turn $t$. 
For text, audio and speaker, we consider context window of turn $t$ to $t-w_t$, turn $t$ to $t-w_a$ and turn $t$ to $t-w_s$, respectively for both the speakers. Note that, here the utterance for the current turn $t$ is also used as a input. 
However for emotion, we consider the context window of turn $t-1$ to $t-w_e$. Here the emotions for the current turn $t$ is not used as a input as we are predicting for the current utterance emotion. 

As we are combining the utterances in turns to capture the inter-dependencies, the number of training examples will reduce significantly. 
We hence restore to light weight machine learning architecture based on 
\cite{VartaRasa2023}, 
where all these multi-modal embeddings, thus created, are concatenated together and passed to the ensemble learning model. 
The concatenated vectors which is the output of the Ensemble model, is again concatenated with the embeddings of the text, audio, emotion and speaker to pass it through another ensemble model to predict the emotions of both the speakers for turn $t$ and forecast the emotions till turn $t+k$. 
As disclosed in the paper \cite{VartaRasa2023}, the light weight architecture will not only make the training feasible for the small number of training examples, but also will help in reducing the time complexity and memory complexity as compared to the deep learning based models like bc-LSTM. 
  
\section{Experiments and Results}
\label{sec:Experiments}
The implementation and the exhaustive experimentation of the proposed ERFC architecture are presented in the following sub-sections.   
Through these results we demonstrate the feasibility of the ERFC for the real-world application. 
\subsection{Experimental Setting}

To train and evaluate the proposed ERFC architecture, we used the Interactive Emotional Dyadic Motion Capture (IEMOCAP) dataset \cite{Busso2008}. 
The data collection comprises approximately 12 hours of audio, transcriptions, video, and motion-capture recordings from five sessions. 
We have targeted 6 emotions (Happy, Excited, Sad, Neutral, Angry and Frustrated) for prediction and forecasting. 
As in most of the literature using IEMOCAP dataset \cite{hazarika2018b, VartaRasa2023}, we also used the first four sessions of transcripts from the IEMOCAP dataset as the training set and the last session as the test set. 
From the available modalities, we use only audio and text from the IEMOCAP dataset for our experiments. 
This is because we are focusing on call center analytics application, where the voice calls are recorded and transcribed using automatic speech recognition (ASR) model to make the real time transcripts available as text. 

Here, for production deployment a turn detection using speaker diarization, where each turn has a part of the speech that belongs to both the speakers.
However, in case of IEMOCAP, the transcript as well as speaker information for each utterances are already available and the same has been used in these experiments.  
For a single speaker in a turn, we may have several distinct utterances.
Utterances of the individual speaker in each turn are concatenated together, and then embeddings are computed on the concatenated utterances. 
The embeddings and emotion labels were created at each \emph{speaker-turn} combination. 
Following are the details on the features creation using embeddings and labels for the multi-modal data:  
\begin{itemize}
\item Text Embeddings: Pre-trained BERT models from the Hugging-face transformers library has been used to extract text embeddings of size 768. 
\item Audio Embeddings: We have extracted the audio features from voice data at a 30 Hz frame rate and a sliding window of 100 ms. The openSMILE has been used to extract 6373 audio features which was further reduced to 250 using Principle Component Analysis (PCA). 
\item Speaker Embeddings: Using the audio data, we have created Speaker embeddings of the size 512 by using Pyannote which was further reduced to 256 using PCA. 
\item Emotions Label: Label encoding has been used for assigning the emotion labels. 
\end{itemize}

\begin{figure}[h]
	\centering
	\includegraphics[scale=0.35]{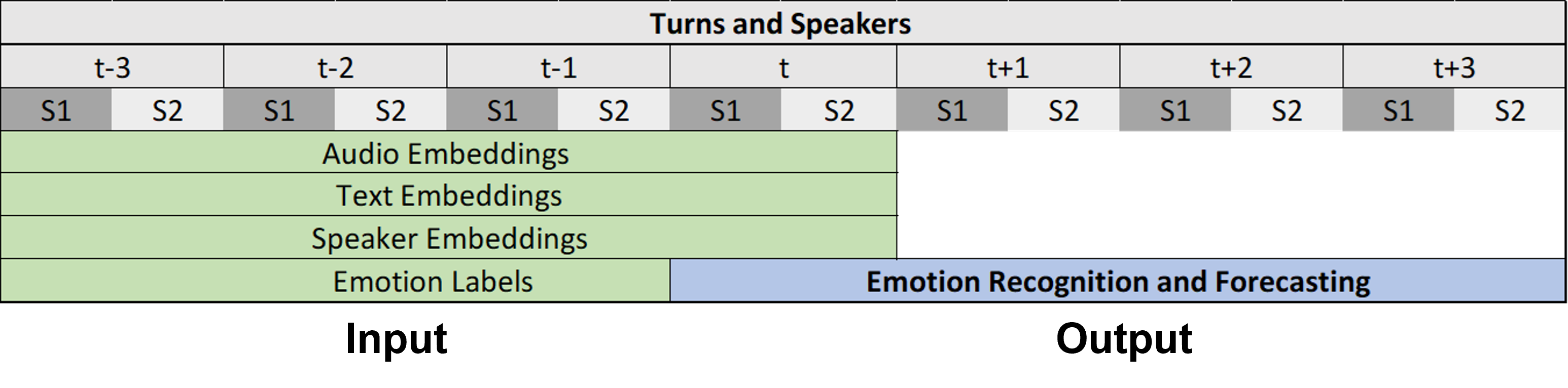}
	\caption{Input output data in the proposed Emotion Recognition and Forecasting in Conversation (ERFC)}
	\label{fig:erfc_io}
\end{figure}

There were $5810$ training utterances from first four sessions and $1623$ test utterances from last session \cite{hazarika2018b}.
However, in our experimental setting, as we are combining \emph{utterances} of both the speakers in \emph{turns}, the number of training and test examples are significantly reduced. 
In this case, we have 1572 training examples and 486 test examples. In a single training example, input features consist of concatenated embeddings of text, audio and speaker along with emotion labels (with attributal values), for their respective contextual window. 
For our experiments, the context widow (number of turns) has been taken as same for all the modalities i.e. $w_t=w_a=w_s=w_e=w$. 
The output label is the emotion for the current utterance and the future utterance for the forecasting horizon of  $k$ turns. 
In our experiments, forecasting horizon $k$ has been taken as $k=3$. 
One representative example to illustrate the data-frame used to train the model for one of the experiments with context window $w=3$ and forecasting horizon $k=3$ is presented in Fig. \ref{fig:erfc_io}.  
Several experiments were carried out as defined in TABLE \ref{experiment}. 
 
\begin{table}[h]
	\centering
	\begin{center}
	\caption{Details of different experiments. E1 - with AVD and no context; E2 - with AVD and 1 turn context; E3 - with AVD and 2 turn context; E4 - with AVD and 3 turn context; E5 - without AVD and 3 turn context; E6 - reduced to 4 emotions by merging \emph{Happy} with \emph{Excited} and \emph{Angry} with \emph{Frustrated} emotions, with AVD and 3 turn context}
\label{experiment}
	\begin{tabular}{|p{2cm}|p{1.5cm}|p{1.5cm}|p{1.5cm}|}
		\hline
		\textbf{Experiment}  & \textbf{Number of Emotions} & \textbf{Emotion Attributes AVD} & \textbf{Context Window Size for all modalities} \\
		\hline
		E1   & 6 & Yes & 0 \\
		\hline
		E2 & 6 & Yes & 1 \\
		\hline
		E3 & 6 & Yes & 2 \\
		\hline
		E4  & 6 & Yes & 3 \\
		\hline
		E5 & 6 & No & 3 \\ 		
		\hline
		E6 & 4  & Yes & 3 \\
		\hline
\end{tabular}
\end{center}
\end{table}

\subsection{Results}
As our focus is on ERFC, our experimental setting is drastically different than the literature \cite{ICON, bcLSTM, DialogueRNN, DialogueGCN, hazarika2018b, VartaRasa2023} which uses the same dataset IEMOCAP, as used in ERC; hence we would not be comparing the results one-on-one with the past literature. 
\subsubsection{Feasibility of Emotion Recognition and Forecasting in Conversation (ERFC)}
For the first time, we showcase emotion forecasting along with emotion recognition in conversation which would be very useful for call center conversation or any other similar applications leveraging emotions from conversations. 
In Experiments E1 to E4, we show results for emotion recognition and emotion forecasting for 6 emotion classes (Happy, Sad, Neutral, Angry, Excited, and Frustrated), when AVD data is used as input, with varying context window size from no context ($w=0$) to context window ($w=3$). 
Here a context window is measured in turns (and not in utterance) where each turn will have atleast one utterance from both the speakers in the two-speaker conversation. 
To reiterate, this will also provide features from both the speakers to utilize the inter-dependencies between the two speakers.     
Table \ref{expt1-4} presents average test accuracies for Experiments E1 to E4  for current turn emotion recognition as well as future turn emotion forecasting. We also present the validation accuracy which reflect that models were trained appropriately. 
As the lags are increased from 0 to 3, we observe that the prediction of emotion for current turn becomes more and more accurate. 
However the same trend is not observed for emotion forecasting for the future turns. 
For the forecasting of emotions, the presence of context upto $one-turn$ was seen to result in better accuracy in comparison to no-context. However, with further increase in context window, there is no further increase in the accuracy. 
\begin{table}[h]
\centering
\caption{Accuracy results for ERFC with respect to different context windows while emotion attributes (AVD) information was also used as actual}
\label{expt1-4}
\begin{tabular}{p{2cm}p{1.2cm}p{1.2cm}p{1.2cm}p{1.2cm}} 
\hline
& \multicolumn{4}{c}{\textbf{Accuracy}} \\
\cline{2-5}
 & \textbf{Validation}  & \multicolumn {3}{c}{\textbf{Test}}\\
\cline{3-5}
\textbf{Experiment} & \textbf{} & \textbf{current turn} & \textbf{future turns average} & \textbf{Average overall} \\ 
\hline
E1 		   & 70.2 & 63.2 & 57.0 & 58.6 \\
\hline
E2 & 71.5 & 72.1 & \textbf{62.5} & \textbf{64.9} \\
\hline
E3 & 71.4 & 72.1 & 62.2 & 64.7 \\
\hline
E4 & 71.5 & \textbf{73.2} & 61.2 & 64.2 \\
\hline
\end{tabular}
\end{table}

\begin{figure}[h]
	\centering
	\includegraphics[width=\linewidth]{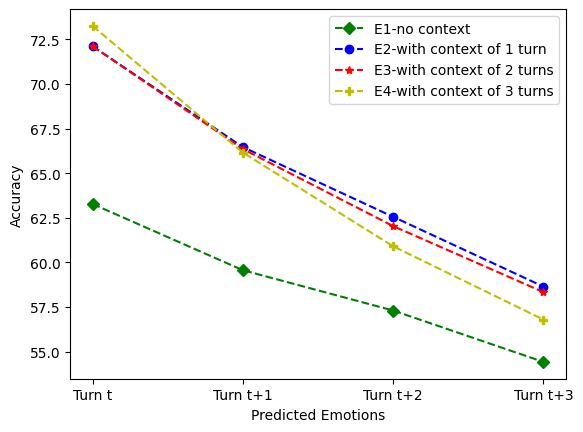}
	\caption{Accuracy for various forecasting horizons for models with different context windows}
	\label{forecasting_horizons}
\end{figure}

In Fig. \ref{forecasting_horizons}, we expand the future turn forecast and show the results with respect to increase in forecasting horizon. As expected, we notice that as we go ahead in forecasting horizon the accuracy is seen to decrease. 
Also this shows clear picture of the significance of the context window over the prediction accuracies for current and future turns. 
While the larger context window $w=3$ is found to be useful for the emotion prediction of the current turn ($t$), it is loosing its relevance for the future turn forecasting $t+3$. 
This is because of the drastic reduction in the training data set for forecasting for the turns much ahead in the horizon as compared to the prediction of the current turn. 
Hence, it is important to note that we can keep the context window dynamic based on the forecasting horizon.

\subsubsection{Impact of Emotion Attributal Features}
Since it is not always possible to have access to AVD features along with emotions, we hence analyse the impact of absence of AVD features in Table \ref{AVD}. Here we compare accuracies for experiments E4 (6 Emotions, $w$=3 and with AVD) with E5 (6 Emotions, $w$=3 and without AVD) and notice the drop in accuracy for E5 as compared to E4 of around 3\% for the prediction of current emotions and around 2\% for the forecast of the future emotions. We hence recommend the use of attributal features if available. 
\begin{table}
\centering
\caption{Impact of emotion attributes (AVD) information on the ERFC prediction accuracies. E4 (with AVD); E5 (without AVD)}
\label{AVD}
\begin{tabular}{p{2cm}p{1.2cm}p{1.2cm}p{1.2cm}p{1.2cm}} 
\hline
& \multicolumn{4}{c}{\textbf{Accuracy}} \\
\cline{2-5}
 & \textbf{Validation}  & \multicolumn {3}{c}{\textbf{Test}}\\
\cline{3-5}
\textbf{Experiment} & \textbf{} & \textbf{current turn} & \textbf{future turns} & \textbf{Average} \\ 
\hline
E4  & 71.5 & \textbf{73.2} & \textbf{61.2} & \textbf{64.2}\\
\hline
E5 & 69.1 & 70.1 & 59.2 & 61.9 \\
\hline
\end{tabular}
\end{table}

\subsubsection{Impact of Number of Emotions}
\begin{figure}[h]
	\centerline{\includegraphics[scale=0.5]{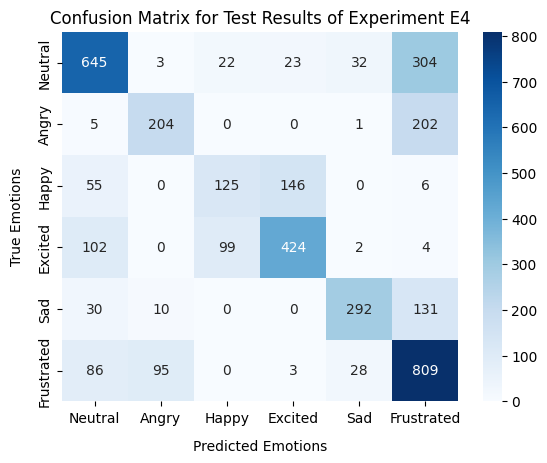}}
	\caption{Confusion matrix for 6 emotions for experiment E4 indicating confusion between different emotional categories}
	\label{confusion}
\end{figure}

In experiment E4, $w=3$ with AVD has been used which shows best result for prediction of current turn emotions. 
The confusion matrix for E4 is shown in Fig. \ref{confusion} which shows the prediction classes distribution. 
This represents $3888$ predictions which are results of predicting for both the speakers for $488$ test examples for $4$ turns (current turn and $3$ turns forecasting horizon).   
The result shows that the model is experiencing confusion between the emotions \emph{happy} and \emph{excited}, as well as \emph{angry} and \emph{frustrated}. 
This is also intuitive as these emotions are similar in nature. 
Hence,  we further experimented (E6) by merging the \emph{happy} and \emph{excited} categories and naming them \emph{happy}, and also merging the \emph{angry} and \emph{frustrated} categories and naming them as \emph{angry} and trained a model for 4 emotion recognition and forecasting from the conversations.
From Table \ref{four_emotions}, it is evident that the overall accuracy for predicting current and future emotions has significantly increased in the case of 4 emotions. 

\begin{table}
\centering
\caption{Effect of number of emotions on ERFC Prediction Accuracies. E4 - 6 emotions; E6- 4 emotions}
\label{four_emotions}
\begin{tabular}{p{2cm}p{1.2cm}p{1.2cm}p{1.2cm}p{1.2cm}} 
\hline
& \multicolumn{4}{c}{\textbf{Accuracy}} \\
\cline{2-5}
 & \textbf{Validation}  & \multicolumn {3}{c}{\textbf{Test}}\\
\cline{3-5}
\textbf{Experiment} & \textbf{} & \textbf{current turn} & \textbf{future turns} & \textbf{Average} \\ 
\hline
E4 & 71.5 & 73.2 & 61.2 & 64.2\\
\hline
E6 & 83.7 & \textbf{82.3} & \textbf{78.3} & \textbf{79.2} \\
\hline
\end{tabular}
\end{table}

\section{Business Impact}
\subsection{New features in Customer Care Support}
In a call center conversation, one of the pressing problems with customer care agents is that how might we pacify the frustration of a customer in a very short time. This requires knowledge of not only what is the real issue but also how frustrated a customer is and if the customer is going to get more angry in the future utterances.  
To answer this, we plan to divide the conversation in different stages such as - Greetings, Problem Discovery, Problem Resolution, Assurance of the Solution and Concluding the call on a positive note. Here in each stage, a challenge in front of agent is to make the customer feel that they are valued. 
While the conversation is moving from one stage to the other stage there is a very short time to act. 
By using the proposed ERFC solution, the forecasted emotions of the customer for the next few turns will come handy in generating useful insights and recommendations for the customer care agent to modulate their emotions in such a way so as to result in an overall positive experience for the customer. 
This will also lead to an increase in the customer happiness. 

\subsection{Other Potential Applications}
Towards this end, we also envisage a completely new application of health consultation for a patient who has hearing and speech impairments. 
The question in front of any hearing and speech impaired individuals when they would like to consult a physician is how might we express our concerns during the diagnosis. 
The proposed EFRC solution can be applied to this problem as well. 
Here the patient communicates in the sign language, a computer vision based sign language recognition models can be trained to process the sign language to generate the transcripts as well and the emotions. 
Transcripts can be generated using the sign language, however their emotions can be captured using their facial expressions. The transcript and emotions can be displayed to the physician as text.  
A physician can comprehend the transcripts and the emotions and converse back in speech. 
An automatic speech recognition (ASR) model can be used to generate the transcript which can be displayed to the patient or using Generative AI, sign language can as well be generated. 
Now ERFC can be used for emotion recognition and forecasting in conversation. Here, the physician will not only have an access to the current emotion of the patient but also will have emotion forecasts available for few more turns in advance. 
This will enable the physician to take the required action to pacify the emotion of the patient and patient will in turn feel confidant with the right words used by the physician. 

Similarly this can be applied to various other applications like Health care - patient-nurse emotional state counselling, Health Insurance claim; human machine interaction - Emotion aware human robot interactions to help machines respond in a right emotional tone or empathy; and Human Resource Management -  recruitment, retention, exit, \& difficult conversations.   

\section{Conclusion}
\label{sec:Conclusion}
In this paper we have shed a new light in the area of Call-Center Application, wherein we propose Emotion Recognition and Forecasting in Conversation (ERFC) which not only helps in recognizing the emotions for the current utterance but also provides emotion forecasting for the future utterances of the speakers in the conversation. 
The intensive experimental results from the implementation of our novel ERFC solution using the IEMOCAP dataset shows the feasibility of the solution. 
The understanding of emotions of future utterance can bring in more empathy and lead to take corrective measures to pacify the negative emotions of the other speaker. 
The proposed ERFC solution can also be applied to a vast variety of applications where insights of the future emotion of the speakers plays a crucial role. 
\\
\\
{
\textbf{Disclaimer:}
This content is provided for general information purposes and is not intended to be used in place of consultation with our professional advisors. The \emph{ERFC} is the property of Accenture and its affiliates and Accenture be the holder of the copyright or any intellectual property over it. No part of this paper may be reproduced in any manner without the written permission of Accenture. Opinions expressed herein are subject to change without notice. 
}

\bibliographystyle{ACM-Reference-Format}
\bibliography{aaai24}

\end{document}